\documentclass{article}

\usepackage{arxiv}
\pdfoutput=1

\usepackage[utf8]{inputenc} 
\usepackage[T1]{fontenc}    
\usepackage{url}            
\usepackage{booktabs}       
\usepackage{amsfonts}       
\usepackage{nicefrac}       
\usepackage{microtype}      
\usepackage{lipsum}
\usepackage{enumitem}
\usepackage{natbib}
\usepackage{url}            
\usepackage{graphicx}
\usepackage{subfigure}
\usepackage[skins]{tcolorbox}
\usepackage{hyperref}
\usepackage[T1]{fontenc}
\usepackage{epigraph}

\usepackage[export]{adjustbox}
\usepackage{amssymb,mathtools}
\usepackage{multirow}
\usepackage{enumitem}
\usepackage{amsmath}
\usepackage{hyperref}
\hypersetup{
    colorlinks=true,
    linkcolor=blue,
    filecolor=magenta,      
    urlcolor=blue,
    citecolor=blue
}
\newcommand{\CIE }{\textit{CIE}}

\makeatletter
\newcommand{\printfnsymbol}[1]{%
  \textsuperscript{\@fnsymbol{#1}}%
}
\makeatother

\newcommand{\sara}[1]{\comments{\textcolor{blue}{}}}
\newcommand{\nyalleng}[1]{\comments{\textcolor{green}{}}}
\newcommand{\gregory}[1]{\comments{\textcolor{navy}{}}}
\newcommand{\emily}[1]{\comments{\textcolor{navy}{}}}
\newcommand{\samy}[1]{\comments{\textcolor{yellow}{}}}

\title{Characterising Bias in Compressed Models}

\author{
 Sara Hooker \thanks{Equal contribution.}\\
  Google Research \\
  \texttt{shooker@google.com} \\
  \And
 Nyalleng Moorosi \printfnsymbol{1}\\
  Google Research \\
  \texttt{nyalleng@google.com} \\
  \And
 Gregory Clark \\
  Google \\
  \texttt{gregoryclark@google.com} \\
  \AND
  
   Samy Bengio \\
  Google Research \\
  \texttt{bengio@google.com} \\
  \And
 Emily Denton \\
  Google Research \\
  \texttt{dentone@google.com} \\
}

\begin{document}
\maketitle
\begin{abstract}
The popularity and widespread use of pruning and quantization is driven by the severe resource constraints of deploying deep neural networks to environments with strict latency, memory and energy requirements. These techniques achieve high levels of compression with negligible impact on top-line metrics (top-1 and top-5 accuracy). However, overall accuracy hides disproportionately high errors on a small subset of examples; we call this subset Compression Identified Exemplars (\CIE{}). We further establish that for \CIE{} examples, compression amplifies existing algorithmic bias. Pruning disproportionately impacts performance on underrepresented features, which often coincides with considerations of fairness. Given that \CIE{} is a relatively small subset but a great contributor of error in the model, we propose its use as a human-in-the-loop auditing tool to surface a tractable subset of the dataset for further inspection or annotation by a domain expert. We provide qualitative and quantitative support that \CIE{} surfaces the most challenging examples in the data distribution for human-in-the-loop auditing.
\end{abstract}

\section{Introduction}\label{sec:introduction}
Pruning and quantization are widely applied techniques for compressing deep neural networks, often driven by the resource constraints of deploying models to mobile phones or embedded devices \citep{2017Andre,8364435}. To-date, discussion around the relative merits of different compression methods has centered on the trade-off between level of compression and top-line metrics such as top-1 and top-5 accuracy \citep{2020blalock}. Along this dimension, compression techniques are remarkably successful. It is possible to prune the majority of weights \citep{tgale_shooker_2019, evci2019rigging} or heavily quantize the bit representation \citep{2017benoit} with negligible decreases to test-set accuracy.

\begin{figure*}[ht] 
\begin{center}
  \includegraphics[width=0.9\textwidth]{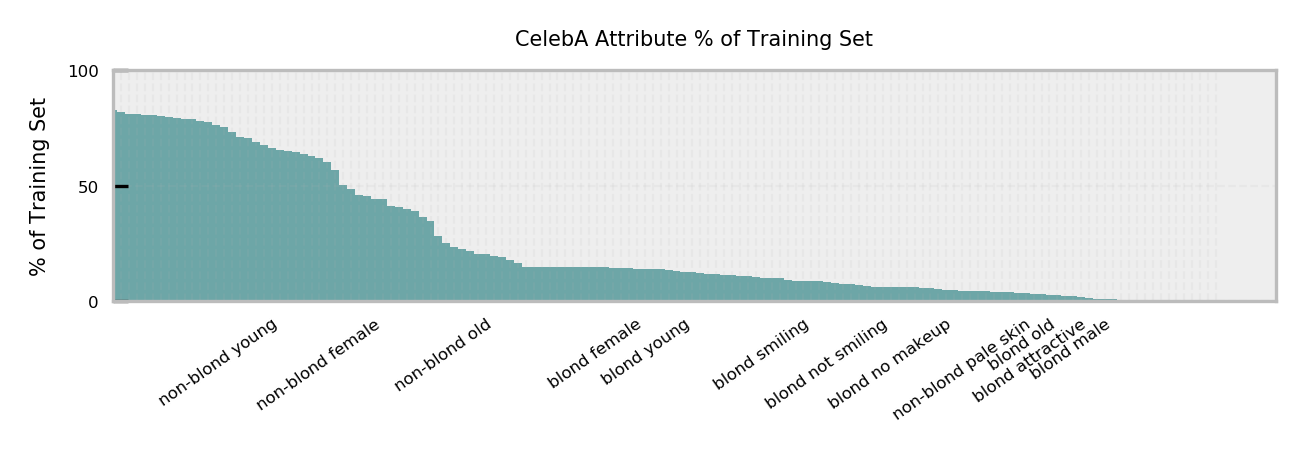}
    \includegraphics[width=0.9\textwidth]{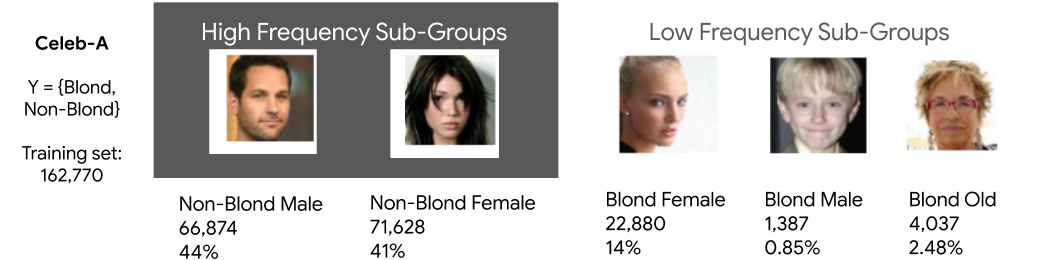}
    \caption{Most natural image datasets exhibit a long-tail distribution with an unequal frequency of attributes in the training data. Below each attribute sub-group in CelebA, we report the share of training set and total frequency count.}
\label{fig:frequency_training_set} \end{center} 
\end{figure*}

However, recent work by \cite{2019shooker} has found that the minimal changes to top-line metrics obscure critical differences in generalization between pruned and non-pruned networks. The authors establish that pruning disproportionately impacts predictive performance on a small subset of the dataset. We build upon this work and focus on the implications of these findings for a dataset with sensitive protected attributes such as gender and age. Our work addresses the question: \textit{Does compression amplify existing algorithmic bias?}

Understanding the relationship between compression and algorithmic bias is particularly urgent given the widespread use of compressed deep neural networks in resource constrained but sensitive domains such as hiring \citep{Dastin_2018, Harwell_2019}, health care diagnostics \citep{2019Hongtao, Gruetzemacher20183DDL, 2019badgeley,2019oakden}, self-driving cars \citep{2017Telsa} and facial recognition software \citep{pmlrbuolamwini18a}. For these tasks, the trade-offs incurred by compression may be intolerable given the impact on human welfare.

We establish consistent results across widely used quantization and pruning techniques and find that compression amplifies algorithmic bias. The minimal changes to overall accuracy hide disproportionately high errors on a small subset of examples. We call this subset Compression Identified Exemplars (\CIE{}). Given two model populations, one compressed and one non-compressed, an example is a \CIE{} if the labels predicted by the compressed population diverges from the labels produced by the non-compressed population.

Reasoning about model behavior is often easier when presented with a subset of data points that is atypical or hard for the model to classify. Our work proposes \CIE{} as a method to surface a tractable subset of the dataset for auditing. One of the biggest bottlenecks for human auditing is the large scale size of modern datasets and the cost of annotating each feature \citep{Veale2017}. For many real-world datasets, labels for protected attributes are not available. In this paper, we show that \CIE{} is able to automatically surface more challenging examples and over-indexes on the protected attributes which are disproportionately impacted by compression. \CIE{} is a powerful unsupervised protocol for auditing. Given that the methodology is agnostic to the presence of attribute labels, \CIE{} allows us to audit multiple attributes all at once. This makes \CIE{} a potentially valuable human-in-the-loop auditing tool for domain experts when labels for underlying attributes are limited.

In Section. \ref{sec:measuring_disparate_impact}, we firstly establish the degree to which model compression amplifies forms of algorithmic bias using traditional error metrics. Section. \ref{sec:auditing_compressed_models} introduces different measures of \CIE{} and motivates the use of \CIE{} as an auditing tool for surfacing these biases when labels are not available for the underlying protected attributes. In Section. \ref{sec:mitigating_harm_compressed_models} we discuss a human-in-the-loop protocol to audit compression induced error.

\section{Characterising Compression Induced Bias in Data with Sensitive Attributes}\label{sec:measuring_disparate_impact}
Recent studies have exposed the prevalence of undesirable biases in machine learning datasets. For example, \cite{buolamwini2018gender} discuss the disparate treatment of darker skin tones due to under-representation within facial analysis datasets, object detection datasets tend to under-represent images from lower income and non-Western regions \citep{shankar2017no, DeVries2019}, activity recognition datasets exhibit stereotype-aligned gender biases \citep{zhao-etal-2017-men}, and word co-occurrences within text datasets frequently reflect social biases relating to gender, race and disability \citep{Garg2017,Hutchinson2020}.

In the absence of fairness-informed interventions, trained models invariably reflect the undesirable biases of the data they are trained on. This can result in higher overall error rates on demographic groups underrepresented across the entire dataset and/or false positive rates and false negative rates that skew in alignment with the over- or under-representation of demographic groups {\it within} a target label.

\begin{figure*}[t] 
\vskip 0.2in 
\begin{center}
  \includegraphics[width=0.9\textwidth]{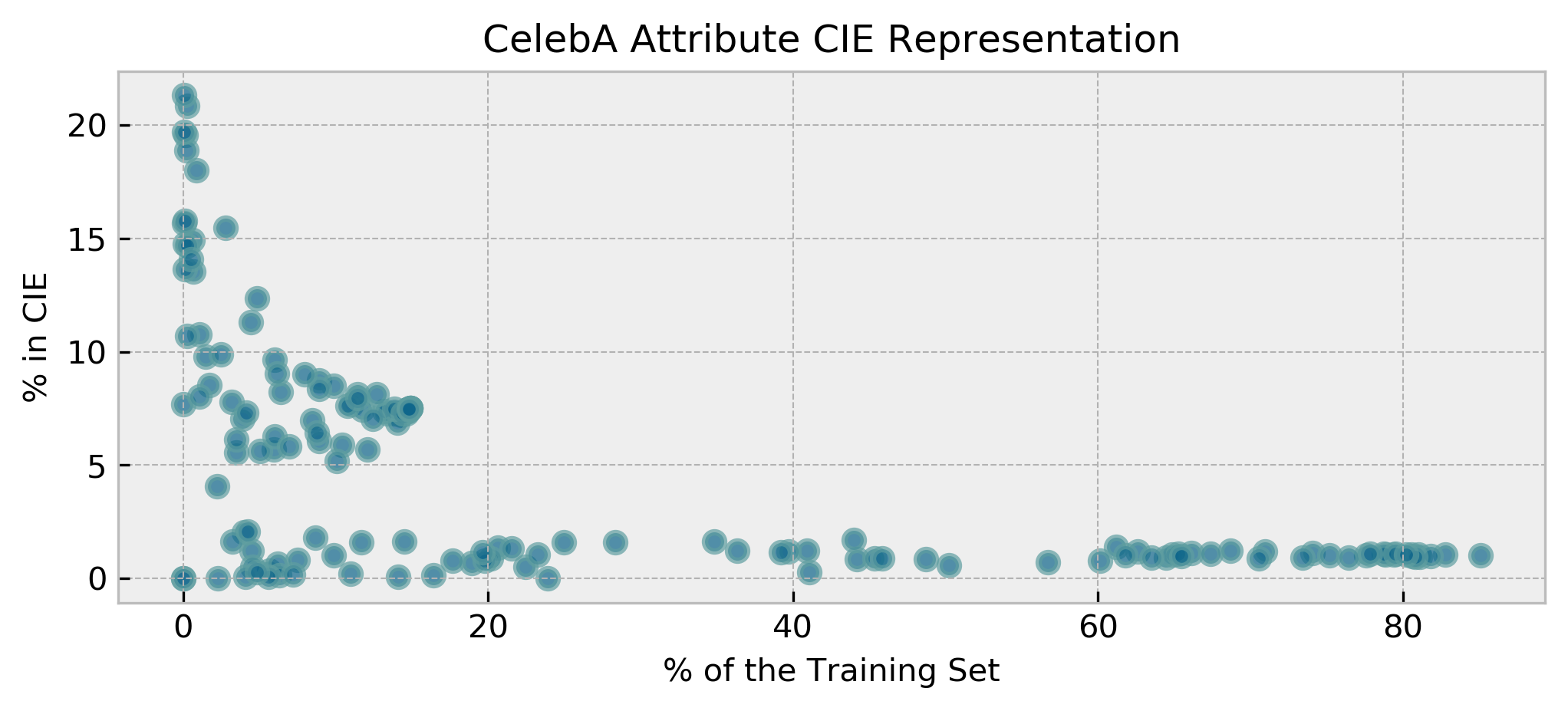}
    \caption{Plot of the fraction of the training set of each attribute in CelebA against the relative representation of each attribute in \CIE{}$_p$. \CIE{}$_p$ over-index on underrepresented attributes in the dataset. In this plot we threshold Taxicab \CIE{} generated from a pruned model at $80 \%$.}
\label{fig:distribution_accuracy} \end{center} 
\end{figure*}

In this section, we firstly establish the degree to which model compression amplifies forms of algorithmic bias using traditional error metrics. Our analysis leverages CelebA \citep{liu2015faceattributes}, a dataset of celebrity faces annotated with 40 binary face attributes and trains a classifier to predict a binary label indicating if the \texttt{Blonde hair} attribute is present. The CelebA dataset is well-suited for our analysis due to the significant correlations between protected demographic groups and the target label (defined by \texttt{Blonde}), as well as the overall under-representation of some demographic groups across the training dataset. As seen in Figure~\ref{fig:frequency_training_set}, CelebA is representative of many natural image datasets where attributes follow a  long-tail distribution \citep{Zhu2014,feldman2019does}. 

\subsection{Methodology} 
Our goal is to understand the implications of compression on model bias and fairness considerations. Thus, we focus attention on two protected unitary attributes \texttt{Male} and \texttt{Young} and one intersectional attribute from the combination of these unitary attributes (i.e \texttt{Young Male}). To characterize the impact of compression on age and gender sub-groups we compare sub-group error rate, false positive rate (FPR) and false negative rate (FNR) between a baseline (i.e. non-compressed) and models pruned and quantized to different levels of compression (i.e. compressed). 

We evaluate three different compression approaches: magnitude pruning \citep{Gupta2017}, fixed point 8-bit quantization \citep{Benoit2017} and hybrid 8-bit quantization with dynamic range \citep{williamson1991dynamically}. In contrast to the pruning which is applied progressively over the course of training, all of the quantization methods we evaluate are implemented post-training. For all experiments, we train a ResNet-18 \citep{He_2015} on CelebA for $10,000$ steps with a batch size of $256$. 

\paragraph{Pruning Protocol} For pruning, we vary the end sparsity for $t \in \{0.3, 0.5, 0.7, 0.9, 0.95, 0.99\}$. For example, $t=0.9$ indicates that $90\%$ of model weights are removed over the course of training, leaving a maximum of $10\%$ non-zero weights at inference time. For the pruning variants, we prune every $500$ steps between $1000$ and $9000$ steps. These hyperparameter choices were based upon a limited grid search which suggested that these particular settings minimized degradation to test-set accuracy across all pruning levels. At the end of training, the final pruned mask is fixed and during inference only the remaining weights contribute to the model prediction. To move beyond anecdotal observations, we train $30$ models for every level of compression considered. Our goal is to have a high level of certainty that differences in predictive performance between compressed and non-compressed models is statistically significant and not due to inherent noise in the stochastic training process of deep neural networks.

\begin{table}[t]
\begin{center}
\begin{tabular}{ccccc}
\toprule
\multicolumn{3}{c}{\textbf{CelebA}} \\
\midrule
\textbf{Fraction Pruned} & \textbf{Top 1}
& \textbf{\# Modal CIEs} \\
\midrule
 0  & 94.73  &  -  \\
0.3 & 94.75 &  555 \\
0.5 & 94.81  &  638 \\ 
0.7 & 94.44 &  990 \\
0.9 & 94.07 &   3229\\
0.95 & 93.39 &  5057\\
0.99 & 90.98 &  8754\\
\midrule
\textbf{Quantization} & \textbf{Top 1}
& \textbf{\# Modal CIEs} \\
hybrid int8 & 94.65 & 404 \\
fixed-point int8 & 94.65 & 414 \\
\bottomrule
\end{tabular}
 \end{center}
 \caption{CelebA top-1 accuracy at all levels of pruning, averaged over runs. The task we consider for CelebA is a binary classification method. We consider exemplar level divergence and classify Compression Identified Exemplars as the examples where the modal label differs between a population of $30$ compressed and non-compressed models.  Note that the CelebA task is a binary classification task to predict whether the celebrity is blond or non-blond. Thus, there are only two classes. 
 ***Note that the number of Taxicab CIEs are just the fraction of the threshold -ie if we threshold at 90\% then the number of \CIE{}s will be 10\% of the dataset. }
 \label{table:appendix_celeba_summary} 
\end{table}

\paragraph{Quantization Protocol} We use two types of post-training quantization.
The first type uses a hybrid ''dynamic range`` approach with 8-bit weights \citep{Alvarez_2016}. The second type uses fixed-point only 8-bit weights  \citep{Vanhoucke_2011, Jacob_2018}, with the first 100 training examples of each dataset as representative examples.  Each of these quantization methods has open source code available.  We use the MLIR implementation via TensorFlow Lite \citep{Jacob_2018, lattner2020mlir}.

\subsection{Results} 

Our baseline non-compressed model obtains $94.73\%$ mean top-1 test-set accuracy (top-5 accuracy is not salient here as it is a binary classification task). Table~\ref{table:absolute_stats} (top row) shows baseline error metrics across unitary and intersectional subgroups. There is a very narrow range of difference in overall test-set accuracy between this baseline and the different compression levels we consider. For example, after pruning $90 \%$ and $95 \%$ of network weights the top-1 test-set accuracy is $94.07 \%$ and $93.39 \%$ respectively. Table \ref{table:compression_test_set_accuracy} provides details of performance at all compression levels for both pruning and quantization.

\begin{table*}[t]
\centering
\begin{tabular}{cccccc}
\toprule
\multicolumn{6}{c}{\textbf{CelebA Top-1 Accuracy}} \\
\midrule    
& \multicolumn{1}{c}{\textbf{Modal \CIE{}s}} & & \multicolumn{3}{c}{\textbf{Taxicab \CIE{}s}} \\
\midrule    
\textbf{Fraction Pruned} &  \textbf{CIEs} & \textbf{All} &  \textbf{90th} &   \textbf{95th} & \textbf{99th} \\
\midrule
           30.0  &                        49.82 &                                 94.75  & 63.58 & 58.49 & 55.35  \\
           50.0  &                        50.55 &                                  94.81  & 63.06 & 58.88 & 54.44   \\
           70.0  &                        52.61 &                                  94.44  & 64.08 & 61.36 & 55.29  \\
           90.0  &                        50.41 &                                  94.07  & 62.35 & 56.60 & 50.10 \\
           95.0  &                        45.57 &                                  93.39  & 60.53 & 51.99 & 43.43  \\
           99.0  &                        39.84 &                                  90.98  & 49.93 & 39.75 & 29.21  \\
\midrule
\textbf{Quantization} &
\\
\midrule
     hybrid int8 &                        48.90 &                                  94.65  & 61.69 & 54.89 & 45.65 \\
fixed-point int8 &                        48.13 &                                  94.65  & 61.68 & 54.41 & 45.15 \\ 
\bottomrule
\end{tabular}
\caption{A comparison of model performance on Compression Identified Exemplars (CIE) relative to performance on the test-set and a sample excluding \CIE{}s (non-\CIE{}s). Evaluation on \CIE{} images alone yields substantially lower top-1 accuracy. Note that CelebA top-5 is not included as it is a binary classification problem.}
\label{table:compression_test_set_accuracy}
\end{table*}

\textit{How does compression amplify existing model bias?}  We find that compression consistently amplifies the disparate treatment of underrepresented protected subgroups for all levels of compression that we consider. While aggregate performance metrics are only minimally affected by compression -- albeit with FNR being amplified to a greater extent that FPR -- we clearly see the newly introduced errors are unevenly distributed across sub-groups. For example,the middle row of Table~\ref{table:relative_stats} shows that at 95\% pruning FPR for \texttt{Male} has a normalized increase of $49.54 \%$ relative to baseline. In contrast, there is far more minimal impact on \texttt{not Male} with a normalized relative increase of only $6.32 \%$. This is less than the overall change in  FPR ($12.72 \%$). We note that this appears closely tied to the overall representation in the dataset, with \texttt{Blond not Male} constituting $14 \%$ of the training set versus \texttt{Blond Male} with only $0.85 \%$. Compression cannibalizes performance on low-frequency attributes in order to preserve overall performance. In Table~\ref{table:subgroup_sparsity} we show that higher levels of compression only further compound this disparate treatment.

\begin{table*}[t]
\begin{center}
\begin{small}
\begin{tabular}{c|c|c|cccc|cccc}
\toprule
\multicolumn{3}{c}{} & \multicolumn{4}{c}{\footnotesize{Unitary}}  & \multicolumn{4}{c}{\footnotesize{Intersectional}}  \\
Model & Metric & Aggregate & M & F & Y & O & MY & MO & FY & FO \\
 \midrule
  \midrule
Baseline & Error & 5.30\% & 2.37\% & 7.15\% & 5.17\% & 5.73\% & 2.28\% & 2.50\% & 5.17\% & 5.73\%  \\
(0\% pruning) & FPR & 2.73\% & 0.93\% & 4.12\% & 2.59\% & 3.18\% & 0.81\% & 1.12\% & 2.59\% & 3.18\%  \\
& FNR & 22.03\% & 62.65\% & 19.09\% & 21.35\% & 24.47\% & 60.45\% & 66.87\% & 21.35\% & 24.47\% \\ 
 \midrule
 \multicolumn{11}{c}{Normalized Difference Between \textbf{1)} Compressed and \textbf{2)} Non-Compressed Baseline} \\
 \midrule
 Compressed & Error & 24.63\% & 24.49\% & 24.67\% & 20.64\% & 35.84\% & 7.96\% & 49.12\% & 20.64\% & 35.84\% \\
(95\% pruning) & FPR & 12.72\% & 49.54\% & 6.32\% & 3.35\% & 36.02\% & 5.37\% & 101.88\% & 3.35\% & 36.02\%  \\
& FNR & 34.22\% & 8.41\% & 40.30\% & 33.83\% & 35.39\% & 9.21\% & 6.98\% & 33.83\% & 35.39\% \\
 \midrule

\bottomrule
\end{tabular}
\end{small}
 \end{center}
 \caption{Performance metrics disaggregated across \texttt{Male} (M), \texttt{not Male} (F), \texttt{Young} (Y), and \texttt{not Young} (O) sub-groups. For all error rates reported, we average performance over $10$ models. \textbf{Top Row}: Baseline error rates, \textbf{Bottom Row:} Relative change in error rate between baseline models and models pruned to  95\% sparsity,}
 \label{table:relative_stats} \end{table*}
 
\section{Auditing Compressed Models in Limited Annotation Regimes}\label{sec:auditing_compressed_models} 

In the previous section, we established that compressed models amplify existing bias using traditional error metrics. However, the auditing process we used and conclusions we have drawn required the presence of labels for protected attributes. The availability of labels is often highly infeasible in real-world settings \citep{Veale2017} because of the cost of data acquisition and privacy concerns associated with annotating protected attributes. In this section, we propose Compression Identified Exemplars (\CIE{}s) as an auditing tool to surface a tractable subset of the data for further inspection or annotation by a domain expert. Identifying a small sample of examples that merit further human-in-the-loop annotation is often critical given the large scale size of modern datasets. \CIE{}s are where the predictive behavior diverges between a population of independently trained compressed and non-compressed models. 

\begin{figure}
\setlength\tabcolsep{3pt}
\begin{tabular}{cccccccc}
\toprule
\multicolumn{7}{c}{\fontfamily{bch}\selectfont \textbf{Blonde}} \\
\midrule
\includegraphics[valign=m,width=0.9in]{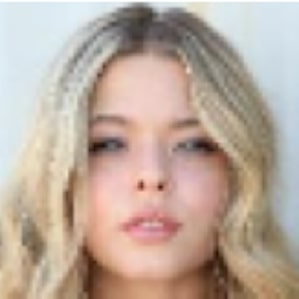} &
 \includegraphics[valign=m,width=0.9in]{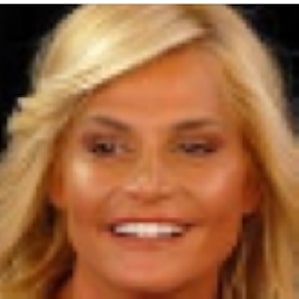} &
  \includegraphics[valign=m,width=0.9in]{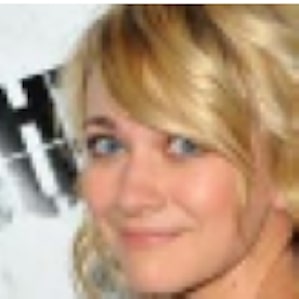}  &
  &
   \includegraphics[valign=m,width=0.9in]{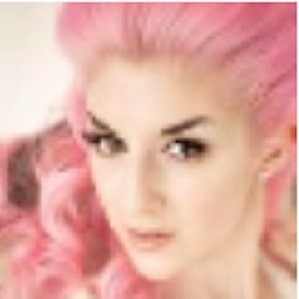} &
 \includegraphics[valign=m,width=0.9in]{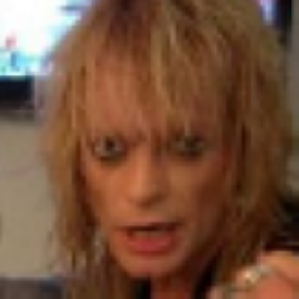} &
   \includegraphics[valign=m,width=0.9in]{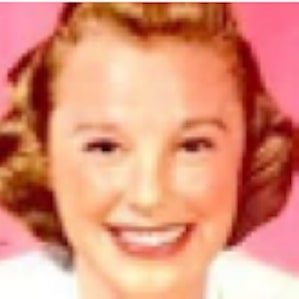} \\
    \addlinespace
    {\fontfamily{qag}\selectfont \small Non-CIE} &
 {\fontfamily{qag}\selectfont \small Non-CIE} & 
 {\fontfamily{qag}\selectfont \small Non-CIE} & &
 {\fontfamily{qag}\selectfont \small CIE} & 
 {\fontfamily{qag}\selectfont \small CIE}  & 
 {\fontfamily{qag}\selectfont \small CIE} \\
    \midrule
  \multicolumn{7}{c}{\fontfamily{bch}\selectfont \textbf{Non-Blonde}}  \\ 
\midrule
  \includegraphics[valign=m,width=0.9in]{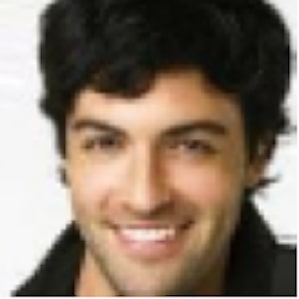} &
 \includegraphics[valign=m,width=0.9in]{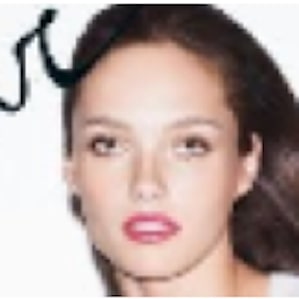} &
  \includegraphics[valign=m,width=0.9in]{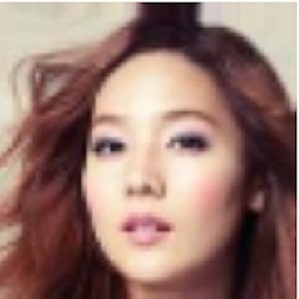} &
  &
 \includegraphics[valign=m,width=0.9in]{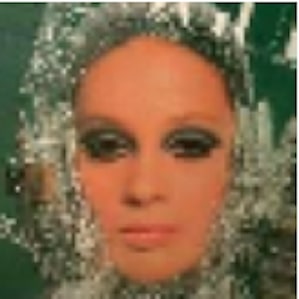} &
 \includegraphics[valign=m,width=0.9in]{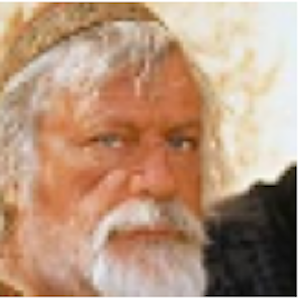} &
  \includegraphics[valign=m,width=0.9in]{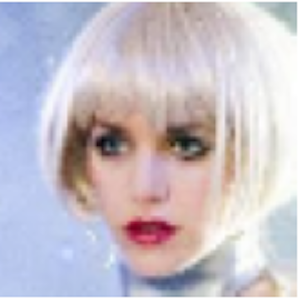} \\
   \addlinespace
{\fontfamily{qag}\selectfont \small Non-CIE} &
 {\fontfamily{qag}\selectfont \small Non-CIE} & 
 {\fontfamily{qag}\selectfont \small Non-CIE} & &
 {\fontfamily{qag}\selectfont \small CIE} & 
 {\fontfamily{qag}\selectfont \small CIE}  & 
 {\fontfamily{qag}\selectfont \small CIE} \\
  \addlinespace
 \end{tabular}
\caption{Compression Identified Exemplars (CIEs) are images where there is a high level of disagreement between the predictions of pruned and non-pruned models. Visualized are a sample of CelebA CIEs alongside a non-CIE image from the same class. Above each image pair is the true label. We train a ResNet-18 on CelebA to predict a binary task of whether the hair color is blond or non-blond.} \label{fig:pie_exemplars_celeba}
\end{figure}

\begin{table*}[ht]
    \centering
    \begin{tabular}{c|c|cc|c}
    \toprule
    \textbf{Metric} &  \textbf{Aggregate} & \multicolumn{2}{c}{ \textbf{Unitary sub-groups}} &  \textbf{Intersectional sub-groups} \\
     \midrule
     \midrule
     \textbf{Error} & \includegraphics[width=0.2\textwidth]{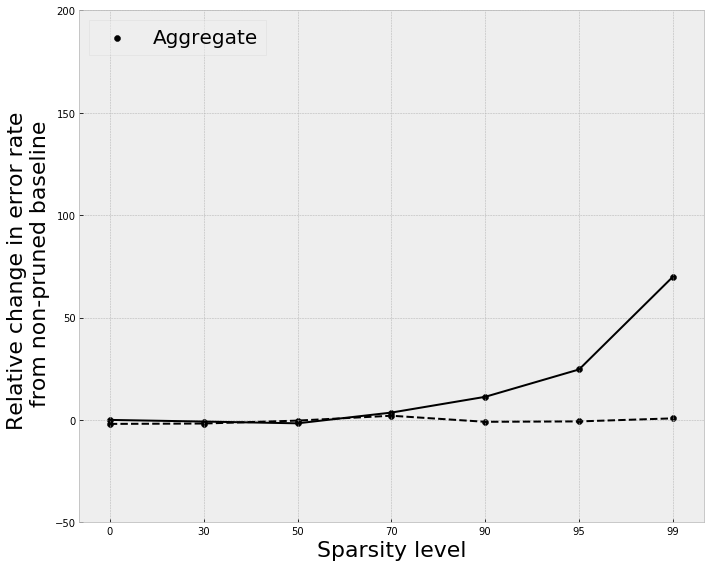} &
    \includegraphics[width=0.2\textwidth]{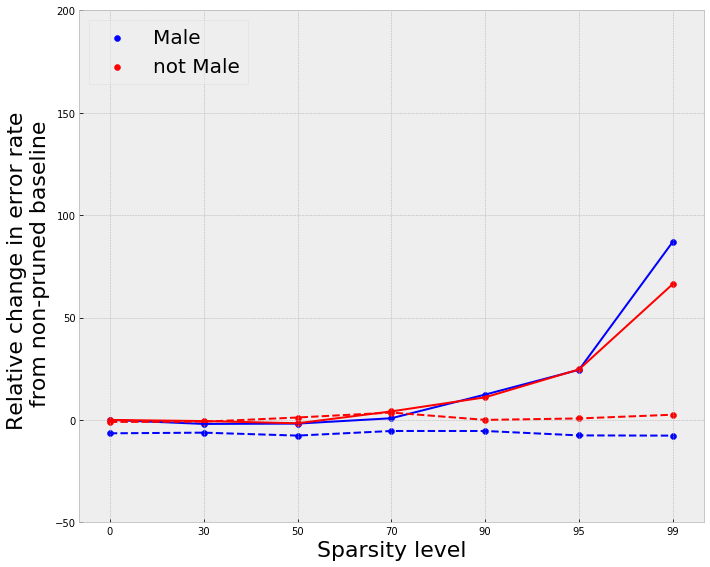} &
    \includegraphics[width=0.2\textwidth]{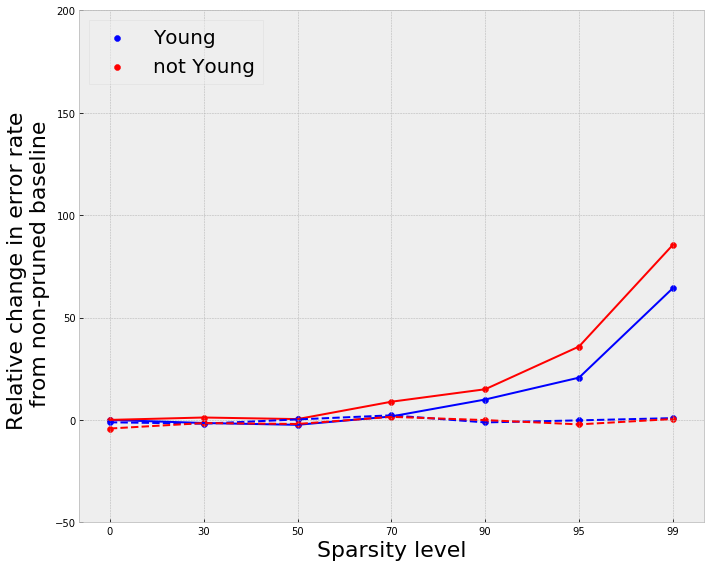} &
    \includegraphics[width=0.2\textwidth]{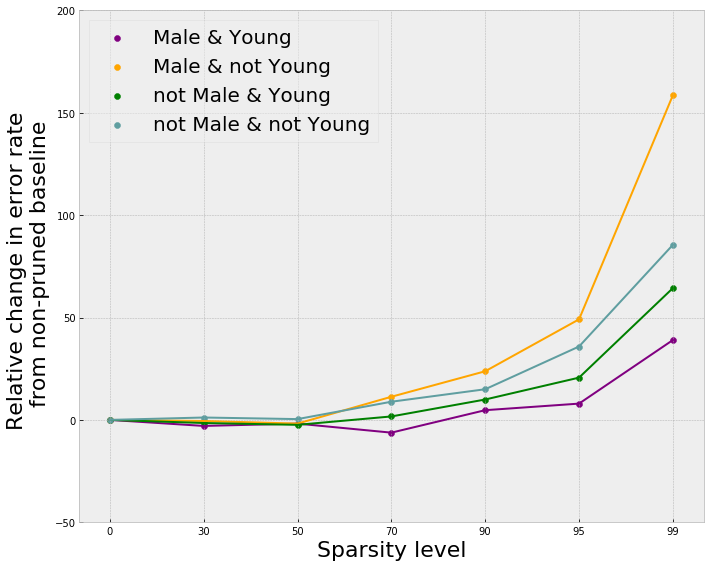} \\
     \midrule
   \textbf{FPR} & \includegraphics[width=0.2\textwidth]{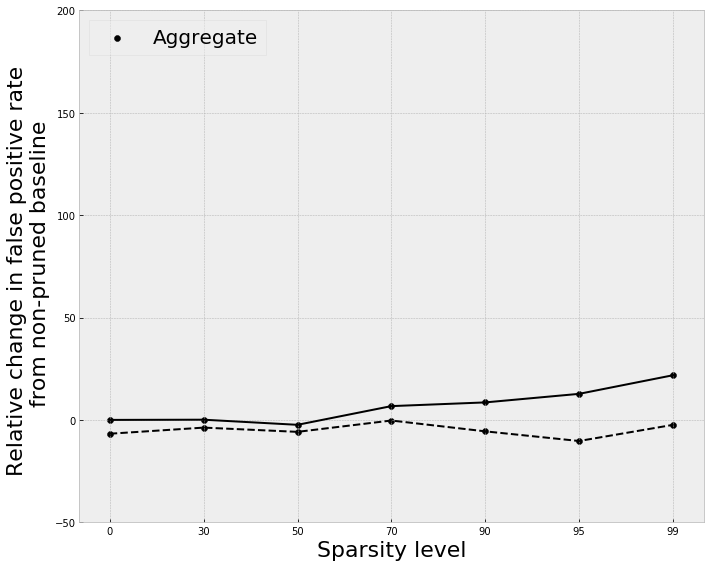} &
    \includegraphics[width=0.2\textwidth]{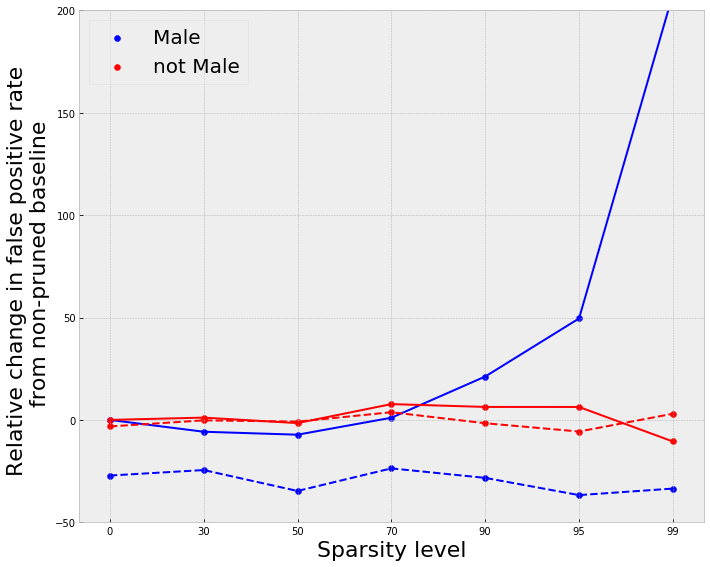} &
    \includegraphics[width=0.2\textwidth]{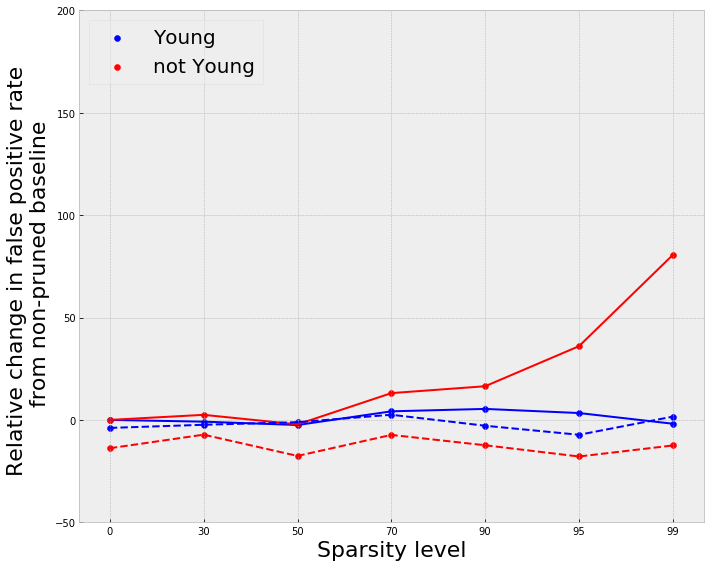} &
    \includegraphics[width=0.2\textwidth]{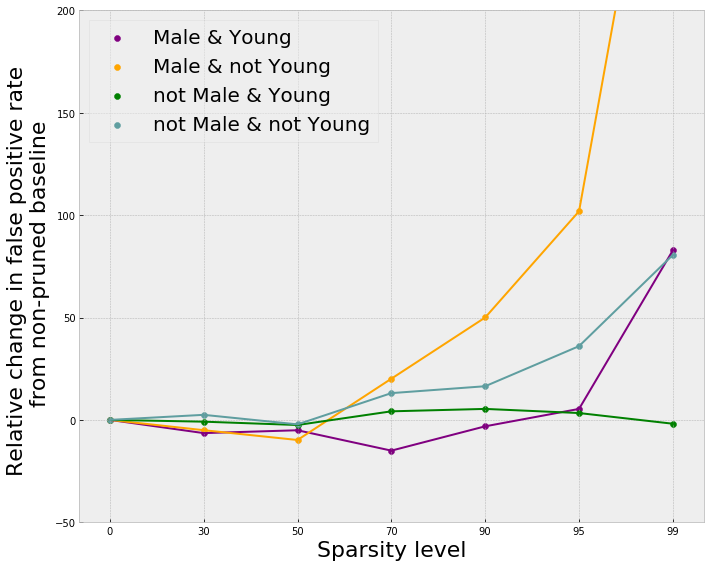} \\
     \midrule
  \textbf{FNR} &  \includegraphics[width=0.2\textwidth]{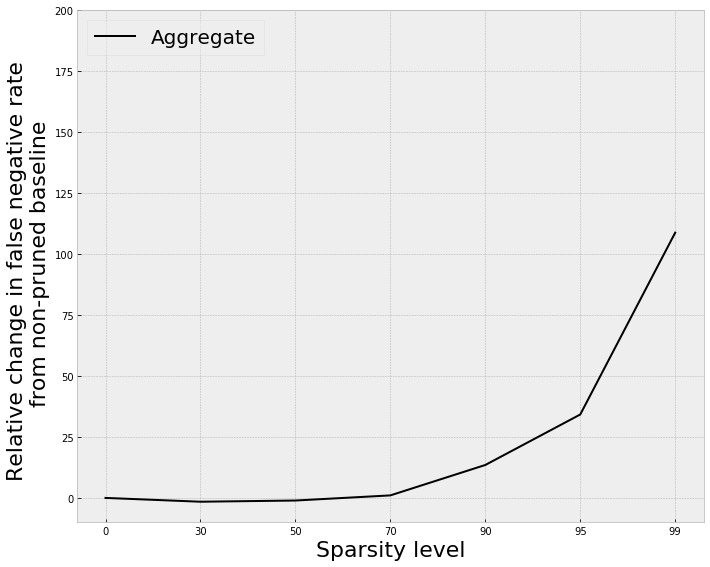} &
    \includegraphics[width=0.2\textwidth]{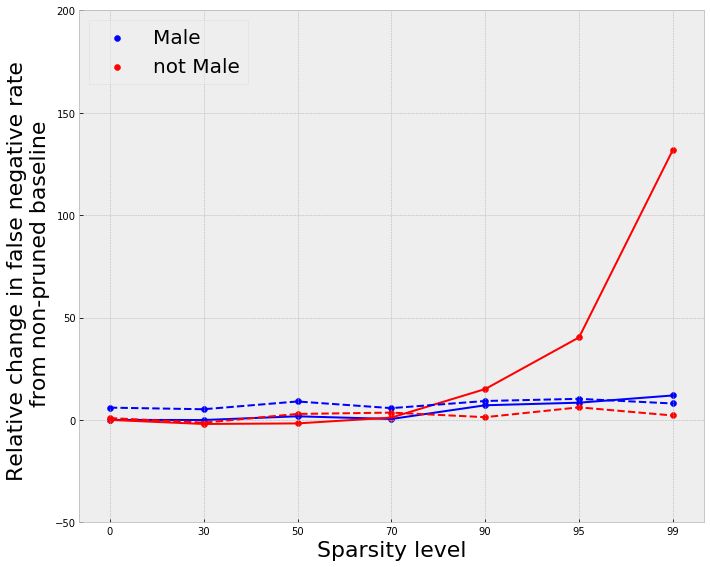} &
    \includegraphics[width=0.2\textwidth]{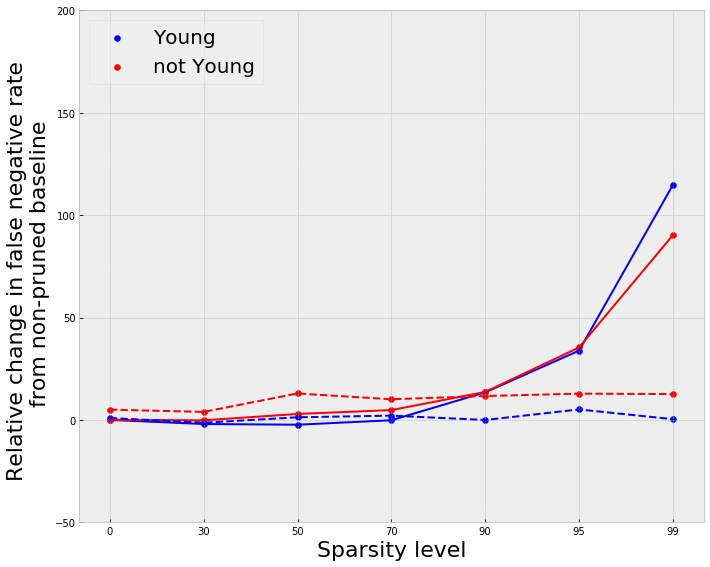} &
    \includegraphics[width=0.2\textwidth]{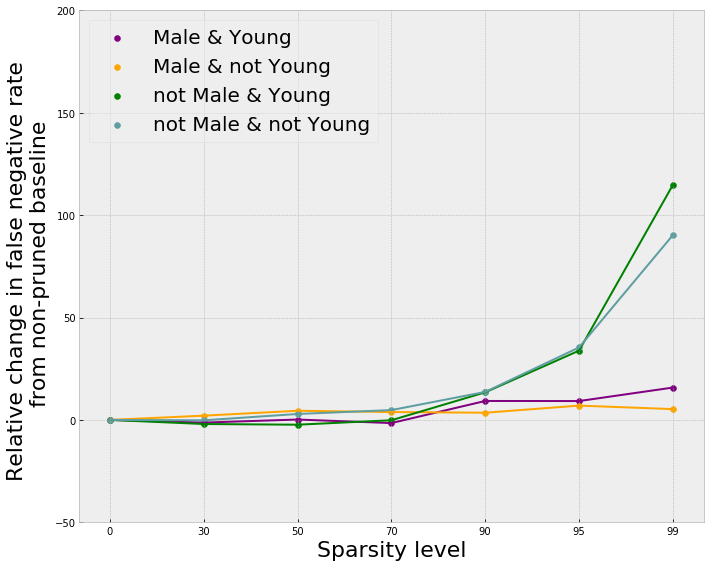} \\
    \midrule
    \end{tabular}
    \caption{For each unitary and intersectional sub-group, we plot the normalized difference of the compressed model, at each level of sparsity (x-axis), relative to the non-compressed model. Note that we threshold the y-axis limit at $100$ for the purposes of standard comparison. \textbf{Top row:} Aggregate error, \textbf{Middle row:} False Positive Rate (FPR), \textbf{Bottom row:} False Negative Rate (FNR)}
    \label{table:subgroup_sparsity}
\end{table*}

\subsection{Divergence Measures}

In additional to the measure of divergence proposed by proposed by \cite{2019shooker} which we term \emph{Modal \CIE{}}, we consider an additional measure of divergence \emph{Taxicab \CIE{}}. We briefly introduce both below. We provide a proof in the appendix of the equivalence of \CIE{}-selection algorithms based on the Jaccard and Taxicab distances.


\paragraph{Modal \CIE{}} \cite{2019shooker}  For set $Y^*_{x,t}$ we find the \textit{modal label}, i.e. the class predicted most frequently by the $t$-compressed model population for exemplar $x$, which we denote $y^M_{x,t}$. Exemplar $x$ is classified as a Modal \CIE{}$_t$ if and only if the modal label is different between the set of $t$-compressed models and the non-compressed models:
\[
\CIE{}_{x,t} = 
\begin{cases}
1 & \text{if $y^M_{x,0} \neq y^M_{x,t}$} \\
0 & \text{otherwise}
\end{cases}
\]

\paragraph{Taxicab \CIE{}} We compute Taxicab distance as the absolute difference between the distribution of labels $y^M_{0}$ from the baseline models and the set $y^M_{t}$ from the compressed models. Given an example $x$, define $B_x = \{b_{x,i}\}$ to be the distribution of labels from a set of baseline models where $b_{x,i}$ is the number of baseline models that label example $x$ with class $i$. Similarly define $V_x = \{v_{x,i}\}$ to be the distribution of labels from a set of variant models where $v_{x,i}$ is the number of variant models that label example $x$ with class $i$.

Let $d_T$ be the Taxicab distance between two label distributions,
\begin{equation*}
    d_T(B_x, V_x) = \sum_i |b_{x,i} - v_{x,i}|.
\end{equation*}


\textbf{Difference between measures proposed} While \emph{Modal} \CIE{} identifies all examples with a changing median label as \CIE{}, \emph{Taxicab} \CIE{} scores the entire dataset allowing for a ranking that can be thresholded by a domain user. Both methods of auditing require no labels for the underlying attributes. That said, note that this turns into a limitation in an overfit $0 \%$ training error regime as without any predictive difference it would not be possible to compute \CIE{} using either measure in the training set.

\subsection{Does ranking by \CIE{} identify more challenging examples?}
\paragraph{Surfacing Challenging Examples \CIE{}} Here, we explore whether \CIE{} divergence measures are able to effectively discriminate between easy and challenging examples. In Table.  \ref{table:compression_test_set_accuracy}, we find that at all levels of compression considered, both \CIE{} metrics surface a subset of data points that are far more challenging for both compressed and non-compressed models to classify. 
For example, while the baseline non-compressed top-1 test set performance on the entire test set is $94.76 \%$, it degrades sharply to $49.82 \%$ and $55.35 \%$ when restricted to Modal \CIE{} (for \CIE{} computed at $t=0.9$) and Taxicab \CIE{} (at percentile $99 \%$) respectively. 
It is hard to compare explicitly the relative difficulty of Modal \CIE{} and Taxicab \CIE{} because the sample sizes are not ensured to be equal. In the appendix, we include the absolute test-set accuracy on a range of Taxicab \CIE{} percentiles and different levels of pruning (Table.). While while examples which are Modal  \CIE{} are more challenging than those identified by Taxicab \CIE{}, for most points of comparison, the results support Taxicab \CIE{} as an effective ranking technique across the \emph{entire} dataset and evidences a monotonic degradation in test-set accuracy as percentile is increased.

\begin{figure}
    \centering
\begin{tabular}{ccc}
\toprule
\multicolumn{2}{c}{\textbf{Top-1 Accuracy on CIE, All Test-Set, Non-CIE}}  \\
\midrule
\includegraphics[width=0.45\textwidth,scale=1]{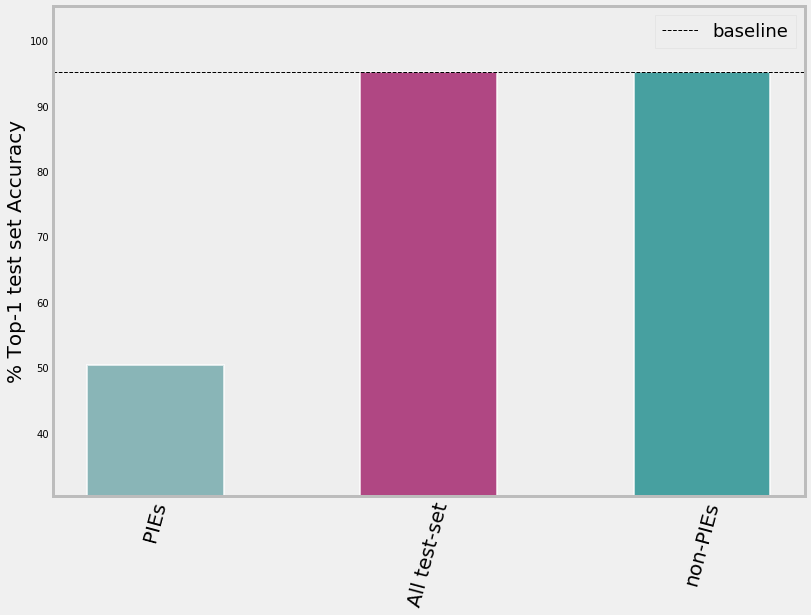} &
 \includegraphics[width=0.45\textwidth]{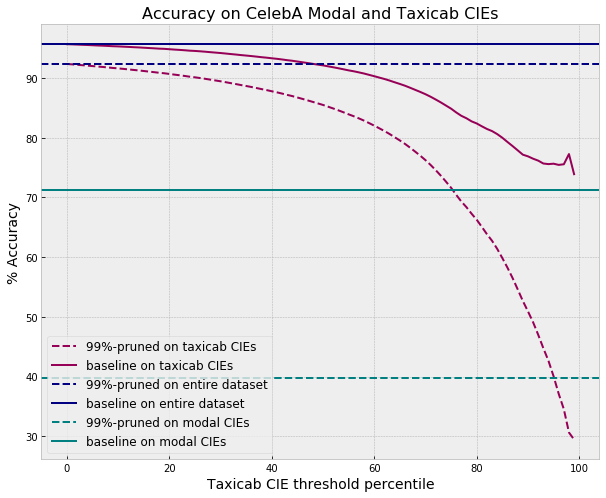} \\
\midrule
\end{tabular}
\caption{\textbf{Right:} A comparison of model performance on $1$) a sample of Modal CIEs against the, $2$) the entire test-set and $3$) a sample excluding CIEs. Evaluation on CIE images alone yields substantially lower top-1 accuracy, \textbf{Left:} Comparison of non-compressed test-set accuracy \textbf{(solid lines)} against compressed $t=99$ pruned test-set accuracy \textbf{(dashed lines)} on \textbf{1)} the entire test-set, with \textbf{2)} Modal \CIE{} identified at $99 \%$ pruning and  \textbf{3)} Taxicab \CIE{} thresholded at different percentiles (x-axis). Any ties for Taxicab \CIE{} are broken at random. Images with high Taxicab \CIE{} scores and or classified as Modal \CIE{} are far more challenging for both the non-compressed and compressed model to classify.}
\label{fig:inference_pie_only}
\end{figure}

\paragraph{Amplified sensitivity of compressed models to \CIE{}} In Fig.~\ref{fig:inference_pie_only}, we plot the test-set accuracy of examples bucketed by Modal \CIE{} and Taxicab \CIE{}.
Overall accuracy drops by less than 3\% between the baseline and pruned models when evaluated on the overall test-set. However, the difference in performance is much larger when we restrict attention to generalization on \CIE{}. Baseline accuracy degrades by $45.86 \%$ on Modal \CIE{} data. For the $99 \%$ pruned model, we see that drop increase to a $52.51 \%$ loss to accuracy. The performance of compressed models degrades far more than non-compressed models on \CIE{}.

\paragraph{Over-indexing of underrepresented attributes on \CIE{}} Here, we ask whether \CIE{} is able to capture the underlying spurious correlation of the target labels with underrepresented attributes. Fairness considerations often coincide with treatment of the long tail. One hypothesis for why compression amplifies bias could be that it impairs model ability to predict accurately on rare and atypical instances. In this experiment, we plot the fraction of the training set of each attribute against the fraction of the attribute in \CIE{}. In Fig.\ref{fig:distribution_accuracy}, we see that underrepresented attributes do indeed over-index on \CIE{}. 

\paragraph{Human-in-the-Loop Auditing with \CIE{}}\label{sec:mitigating_harm_compressed_models} Relying on underlying attribute labels to mitigate the harm of compression is common in fairness literature \cite{Hardt2016}. However, this is costly and hinges on the assumption there has been extensive labelling of all protected attributes. Here, we propose the use of \CIE{} as a human-in-the-loop auditing tool. Through the use of a threshold and Taxicab \CIE{}, a practitioner can select examples the model performs the worst on for an audit. This will surface all examples regardless of attribute label and will therefore allow for an intersectional audit.

\section{Related Work}\label{sec:related_work}
Despite the widespread use of compression techniques, articulating the trade-offs of compression has overwhelming centered on change to overall accuracy for a given level of compression \citep{Strom97sparseconnection, Cun90optimalbrain, evci2019rigging, 2017Narang}. Recent work by \citep{NIPS2018_7308, Sehwag2019} has considered sensitivity of pruned models to a a different notion of robustness: $L^p$ norm adversarial attacks. Our work builds upon recent work by \citep{2019shooker} which measures difference in generalization behavior between compressed and non-compressed models. In contrast to this work, we connect the disparate impact of compression to fairness implications and are interested in \emph{both} characterizing and mitigating the harm. Leveraging a subset of data points to understand model behaviour or to audit a dataset fits into a broader literature that aims to characterize input data points as prototypes -- ``most typical" examples of a class -- \citep{Carlini2018PrototypicalEI, 2020arXiv200811600A, 2017Stock_Cisse,2020arXiv200203206J}) or outside of the training distribution \citep{2016Hendrycks, 2018Masana}.

\section{Conclusion}
We make three main points in this paper. We illustrate that while overall error is largely unchanged when a model is compressed, there is a set of data which bears a disproportionately high portion of the error. We highlight fairness issues which can result from this phenomena by considering the impact of compression on CelebA. Second, we show that this set can be isolated by annotating points where the labels produced by the dense models diverge from the labels from the compressed population. Finally, we propose the use of \CIE{} as an attribute agnostic human-in-the-loop auditing tool.
\clearpage

\bibliography{main}
\bibliographystyle{neurips_2020}

\clearpage
\appendix
\section{Appendix}
\subsection{Equivalence of Taxicab \CIE{} and Jaccard \CIE{}}

In addition to Modal \CIE{} and Taxicab \CIE{}, we considered comparing sets of labels with a weighted Jaccard distance \cite{Chierichetti2010}.
We find that the \CIE{}-selection algorithm based on the Jaccard distance and the algorithm based on the Taxicab distance are equivalent.
In this section, we prove that for two examples $x$ and $y$, Jaccard \CIE{} prefers $x$ over $y$ if and only if Taxicab \CIE{} also prefers $x$ over $y$.

Given an example $x$, define $B_x = \{b_{x,i}\}$ to be the distribution of labels from a set of baseline models
where $b_{x,i}$ is the number of baseline models that label example $x$ with class $i$.
Similarly define $V_x = \{v_{x,i}\}$ to be the distribution of labels from a set of variant models
where $v_{x,i}$ is the number of variant models that label example $x$ with class $i$.


Let $d_T$ be the Taxicab distance between two label distributions,
\begin{equation*}
    d_T(B_x, V_x) = \sum_i |b_{x,i} - v_{x,i}|.
\end{equation*}
Let $d_J$ be the Jaccard distance between two label distributions,
accounting for multiplicity of labels,
\begin{equation*}
    d_J(B_x, V_x) = 1 - \frac{\sum_i \min(b_{x,i}, v_{x,i})}{\sum_i \max(b_{x,i}, v_{x,i})}.
\end{equation*}
First notice that
\begin{equation}
    \max(b, v) - \min(b, v) = |b - v|
    \label{eq:min-max-diff}
\end{equation}
for all integers $b$ and $v$.
Assume that each family contains $N$ models.
Then,
\begin{equation}
    \sum_i \max(b_{x,i}, v_{x,i}) = N + \frac{1}{2}\sum_i |b_{x,i} - v_{x,i}|
    \label{eq:total-label-diff}
\end{equation}
as shown by pairing equal baseline and variant labels with each other and counting the labels that are left over.

\begin{table*}
\begin{center}
\begin{small}
\begin{tabular}{c|c|c|cccc|cccc}
\toprule
\multicolumn{3}{c}{} & \multicolumn{4}{c}{\footnotesize{Unitary}}  & \multicolumn{4}{c}{\footnotesize{Intersectional}}  \\
Model & Metric & Aggregate & M & F & Y & O & MY & MO & FY & FO \\
 \midrule
  \midrule
Baseline & Error & 5.30\% & 2.37\% & 7.15\% & 5.17\% & 5.73\% & 2.28\% & 2.50\% & 5.17\% & 5.73\%  \\
(0\% pruning) & FPR & 2.73\% & 0.93\% & 4.12\% & 2.59\% & 3.18\% & 0.81\% & 1.12\% & 2.59\% & 3.18\%  \\
& FNR & 22.03\% & 62.65\% & 19.09\% & 21.35\% & 24.47\% & 60.45\% & 66.87\% & 21.35\% & 24.47\% \\ 
 \midrule
 \midrule
 Compressed & Error & 6.61\% & 2.95\% & 8.92\% & 6.23\% & 7.78\% & 2.47\% & 3.73\% & 6.23\% & 7.78\%  \\
(95\% pruning) & FPR & 3.08\% & 1.39\% & 4.39\% & 2.67\% & 4.32\% & 0.86\% & 2.25\% & 2.67\% & 4.32\%  \\
& FNR  & 29.57\% & 67.92\% & 26.78\% & 28.57\% & 33.13\% & 66.02\% & 71.53\% & 28.57\% & 33.13\% \\
 \midrule
\bottomrule
\end{tabular}
\end{small}
 \end{center}
 \caption{Absolute performance metrics dis-aggregated across unitary and intersection sub-groups. For all error rates reported, we average performance over $10$ models. \textbf{Top Row}: Baseline error rates, \textbf{Bottom Row:} Error rates of models pruned to  95\% sparsity.}
 \label{table:absolute_stats} \end{table*}

Furthermore, notice that,
\begin{equation}
    s > t \iff \frac{s}{r + s} > \frac{t}{r + t}
    \label{eq:fraction-inequality}
\end{equation}
for all positive real numbers $s, t, r \in \mathbb{R}^+$.
We apply \eqref{eq:min-max-diff}, \eqref{eq:total-label-diff}, and \eqref{eq:fraction-inequality} in order to show the desired equivalence.

\begin{alignat*}{2}
    && d_J(B_x, V_x) &> d_J(B_y, V_y)\\
\iff&& 1 - \frac{\sum_i \min(b_{x,i}, v_{x,i})}{\sum_i \max(b_{x,i}, v_{x,i})} &> 1 - \frac{\sum_i \min(b_{y,i}, v_{y,i})}{\sum_i \max(b_{y,i}, v_{y,i})} \\
\iff&& \frac{\sum_i |b_{x,i} - v_{x,i}|}{\sum_i \max(b_{x,i}, v_{x,i})} &> \frac{\sum_i |b_{y,i} - v_{y,i}|}{\sum_i \max(b_{y,i}, v_{y,i})}\\
\iff&& \frac{\sum_i |b_{x,i} - v_{x,i}|}{2N + \sum_i |b_{x,i} - v_{x,i}|} &> \frac{\sum_i |b_{y,i} - v_{y,i}|}{2N + \sum_i |b_{y,i} - v_{y,i}|}\\
\iff&& \sum_i |b_{x,i} - v_{x,i}| &> \sum_i |b_{y,i} - v_{y,i}|\\
\iff&& d_T(B_x, V_x) &> d_T(B_y, V_y)\\
\end{alignat*}

\subsection{Absolute Performance Metrics Disaggregated}

In Table.\ref{table:absolute_stats}, we include the absolute performance for every sub-group and intersection of sub-group that we consider.

\end{document}